\documentclass[conference]{IEEEtran}
\usepackage{fancyhdr}
\fancypagestyle{firstpage}
{
    \fancyhead[L]{}    
    \fancyhead[R]{Submitted to ICRA 2021, under peer review}
}

\ifCLASSINFOpdf
    \usepackage[pdftex]{graphicx}
\else
\fi

\usepackage{geometry}
\begin{document}
%
\title{Pose Estimation of Specular and Symmetrical Objects}

\author{\IEEEauthorblockN{Jiaming Hu*\IEEEauthorrefmark{2},
Hongyi Ling*, 
Priyam Parashar,
Aayush Naik, and
Henrik Christensen}
\IEEEauthorblockA{\IEEEauthorrefmark{2}jih189@eng.ucsd.edu}
\IEEEauthorblockA{Computer Science and Engineering Department,\\
UC San Diego, La Jolla }}

%


\maketitle

\begin{abstract}
In the robotic industry, specular and textureless metallic components are ubiquitous.  The 6D pose estimation of such objects with only a monocular RGB camera is difficult because of the absence of rich texture features. Furthermore, the appearance of specularity heavily depends on the camera viewpoint and environmental light conditions making traditional methods, like template matching, fail. In the last 30 years, pose estimation of the specular object has been a consistent challenge, and most related works \cite{chang2009specularcues}\cite{liu2011pose}\cite{liu2010multiflash} require massive knowledge modeling effort for light setups, environment, or the object surface. On the other hand, recent works\cite{kehl2017ssd}\cite{kendall2015posenet}\cite{li2018deepim}\cite{park2019pix2pose} exhibit the feasibility of 6D pose estimation on a monocular camera with convolutional neural networks(CNNs) however they mostly use opaque objects for evaluation.
This paper provides a data-driven solution to estimate the 6D pose of specular objects for grasping them, proposes a cost function for handling symmetry, and demonstrates experimental results showing the system's feasibility.
\end{abstract}


%
\IEEEpeerreviewmaketitle

\section{Introduction}
\thispagestyle{firstpage}
In manufacturing, it is common to encounter objects that are specular and have significant symmetries. Even though such objects have high-fidelity data available for describing their geometry, depending upon lighting conditions and object-camera interaction angles they can appear significantly different from baseline expectations. Inherently, such objects flout two most explored inter-object characteristics in pose estimation, i.e. uniqueness in terms of invariant visual features and geometrical asymmetry. Pose estimation for textureless objects with significant symmetries and specular surfaces, which are frequently used in the assembly domain, without a depth camera is an interesting challenge being studied\cite{chang2009specularcues}\cite{liu2011pose}\cite{liu2010multiflash} for 20 years. 

In last 5 year, pose estimation with an RGB camera using convolutional neural networks (CNNs)\cite{xiang2017posecnn}\cite{kehl2017ssd} become popular solution for textureless objects. Later, for better performance, people try to include the model information of the object as the input to the net\cite{li2018deepim}\cite{zakharov2019dpod}, but they require the surface information of the object. For textureless and symmetric objects with specular surfaces, the surface information is hard to collect because the surface of specular objects vary significantly in different environments and viewpoints. Furthermore, the types of symmetry are various like cylindrical, bilateral, and spherical symmetries, so it also requires the pose estimation has to handle them respectively.

This paper proposes a cascaded neural network architecture based 6D pose estimation approach which can deal with symmetric problems and uses readily available model information but without the surface detail of objects. Furthermore, we observed that the usual loss functions used in pure computer vision applications do not work as well when object-symmetries are present, resulting in translation being weighted more heavily than rotation for compensation. Therefore, we propose a new loss function that considers the full 6D pose loss in the training.

The rest of this paper is organized as follows. After discussing the related work in Section 2, we introduce the whole pipeline in Section 3. Implement details are discussed in Section 4 and experiments are presented in Section 5. Section 6 concludes the paper.

\begin{figure}[!t]
\centering
\includegraphics[width=3.5in]{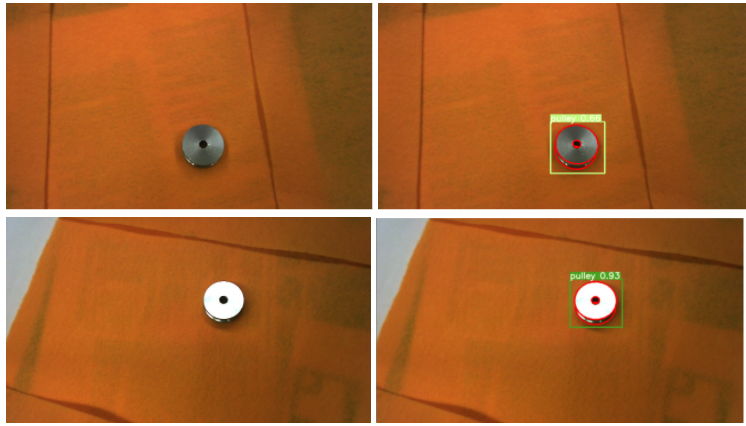}
\caption{Example images of symmetric and specular objects with their predictions.}
\label{rough}
\end{figure}

\begin{figure*}[!t]
\centering
\includegraphics[width=7.0in]{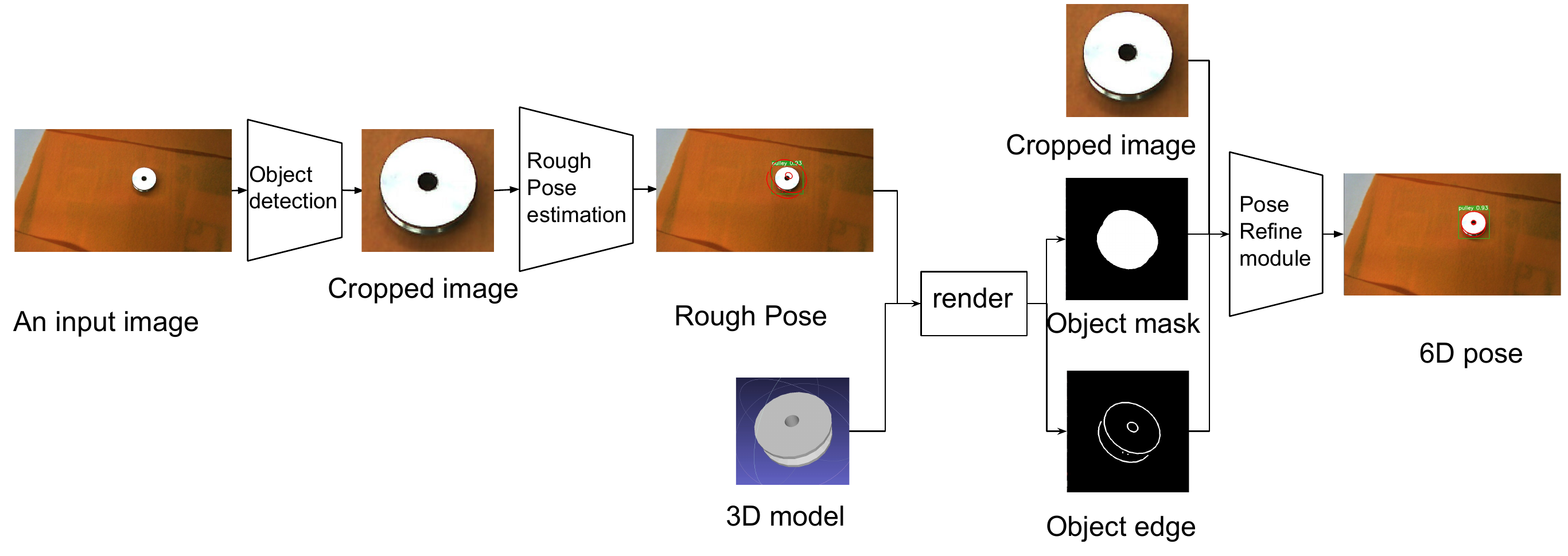}
\caption{Whole pipeline of pose estimation.}
\label{pipeline}
\end{figure*}

\section{Related works}
\subsection{Traditional method}
Classical model-based methods can be categorized into two main schools: keypoint-feature based and template-based. Keypoint-feature based methods \cite{grundmann2010robust} presume geometric and illumination invariance over the keypoints\cite{lowe2004distinctive} and do well for matte objects with colors and textures. However, textureless objects were a problem for this method, therefore template matching was proposed \cite{choi2012robust}\cite{alvarez2013junction} which created a library of edge-based templates for several possible camera-to-object viewpoints. The best aligned template is the most probable viewpoint and pose of the object. Our specular objects, however, do not have clear edges and sometimes can show edges which do not exist due to reflective surfaces. 
VVS(virtual visual servoing)-based pose estimation method \cite{oumer2014monocular} proposes a solution for specular objects by adapting keypoint and edge-based approach with known simple geometry. Besides s\cite{chang2009specularcues,roth2006specular,sankaranarayanan2010specular} use specular flow to deal with such objects. The paper\cite{chang2009specularcues} generates a library of specular flows for 3D poses as illumination invariant features and compares them with specular flow of the scene. While accurate in results such methods require detailed modeling of the environment and it is valuable to explore methods which can use data from models of limited object-sets rather than the possible environment configurations which can be uncountable.

\subsection{Deep learning method}
Recently, deep learning methods\cite{peng2019pvnet}\cite{zakharov2019dpod}\cite{park2019pix2pose}\cite{zhao2018estimating} have been used to detect robust keypoint. Besides there are some holistic approaches like  PoseNet \cite{kendall2015posenet} and PoseCNN \cite{xiang2017posecnn}. PoseNet uses a CNN architecture to directly regress a 6D pose from an RGB image. Directly localizing the objects in 3D space is difficult due to the large search space. Besides, translational and rotational loss should be balanced by tuning a hyper-parameter during training. To overcome these problems, PoseCNN breaks this up in multiple stages. To further simplify the problem other methods have formulated this as a viewpoint classification, ie. \cite{kehl2017ssd}. Discretizing the orientation space and converting the 3D rotation estimation to a classification problem makes it simpler to train a neural network. However, such discretization can only produce a coarse pose estimation due to the number of classes. While limited accuracy is possible with viewpoint based methods, robotics requires high accuracy to facilitate interactions such as grasping of objects. 

Deep learning based pose refinement has been utilized to improve the result of pose estimation in the recent study. Most recent works\cite{li2018deepim} \cite{manhardt2018deep} have a very similar idea of predicting a relative transformation from initial pose to the predicted pose with a mixed input. The input of the network is typically a mixture of raw image samples and rendered images to speed-up data capture. However, rendering objects with specular surfaces poses a challenge as such objects appear significantly different upon light conditions. 

\section{Methodology}
We have adopted a cascaded approach to pose estimation using a coarse to fine strategy, where each stage is realized using a neural network. The benefit of this design is threefold. First, reusing successful work\cite{Redmon2016YOLO} can be possible. The second, it can split one hard task into several easier tasks, so analysis can be easier. The third, it will solve the common problem in pose data collection, and the detail will be discussed in section \ref{sec:datacollection}. The cascaded architecture is shown in figure \ref{pipeline}.

\subsection{Pipeline}
In order to break-down the complexity of this problem, both in terms of training and learning, we designed our pipeline to remove spurious information and add relevant information in stages. The initial stage is an object detection system that returns a smaller cropped image which helps in cutting down on training time based on image-size. Based on \cite{li2018deepim, manhardt2018deep} we subsequently have a two stage rough-to-fine pose model with supplemented data from rendered images.

\subsection{Object detection}
Recently a number of deep learning methods have been recorded for object detection that have superior performance. The reference model is YOLO \cite{Redmon2016YOLO}. A challenge is detection of objects from a limited set of pixels, which is not a challenge here. Therefore, in our study, we choose to use YOLO as the object detection solution.

\subsection{Rough pose estimation}
\subsubsection{Architecture detail}
The rough pose estimator is providing a full pose (rotation and translation) estimate.  For translation estimation, the network computes the offset from image center to 2D object center but not the distance from the camera. For rotation the viewpoint and in-plane rotation is estimated. In simple terms, we homogeneously sample 64 points on the fibonacci sphere around the object, and a vector from one point to the object center will be considered as an orientation. Each view can spin along the direction with some degree, and such an angle is defined as an in-plane rotation. To simplify the problem, the in-plane rotation is discretized to 60 different degrees in our approach.

The rough pose estimator strategy is illustrated in figure \ref{rough}. The cropped image is resized to size 240x240 and ResNet34, which is commonly used as backbone in vision, is used to generate a feature map. In contrast to SSD6D\cite{kehl2017ssd} which predicts object class, bounding box, offset, viewpoint, and in-plane rotation, in our work the resulting representation is utilized in a fully connected layer to estimate only the latter three. With this information and the distance between object and camera, rough pose can be calculated. Similar to SSD6D \cite{kehl2017ssd}, the network does not predict the depth, rather uses size of bounding box to provide rough depth. This decision was taken because we recognize that the depth information is hard to be estimated and training the model together with the depth information causes worse performances on other predictions. Thus, for the depth, we decide to rely on the pose refinement which can use rough depth and approximate the real depth by aligning the predicted object with the real object in image.

\begin{figure}[!t]
\centering
\includegraphics[width=3.5in]{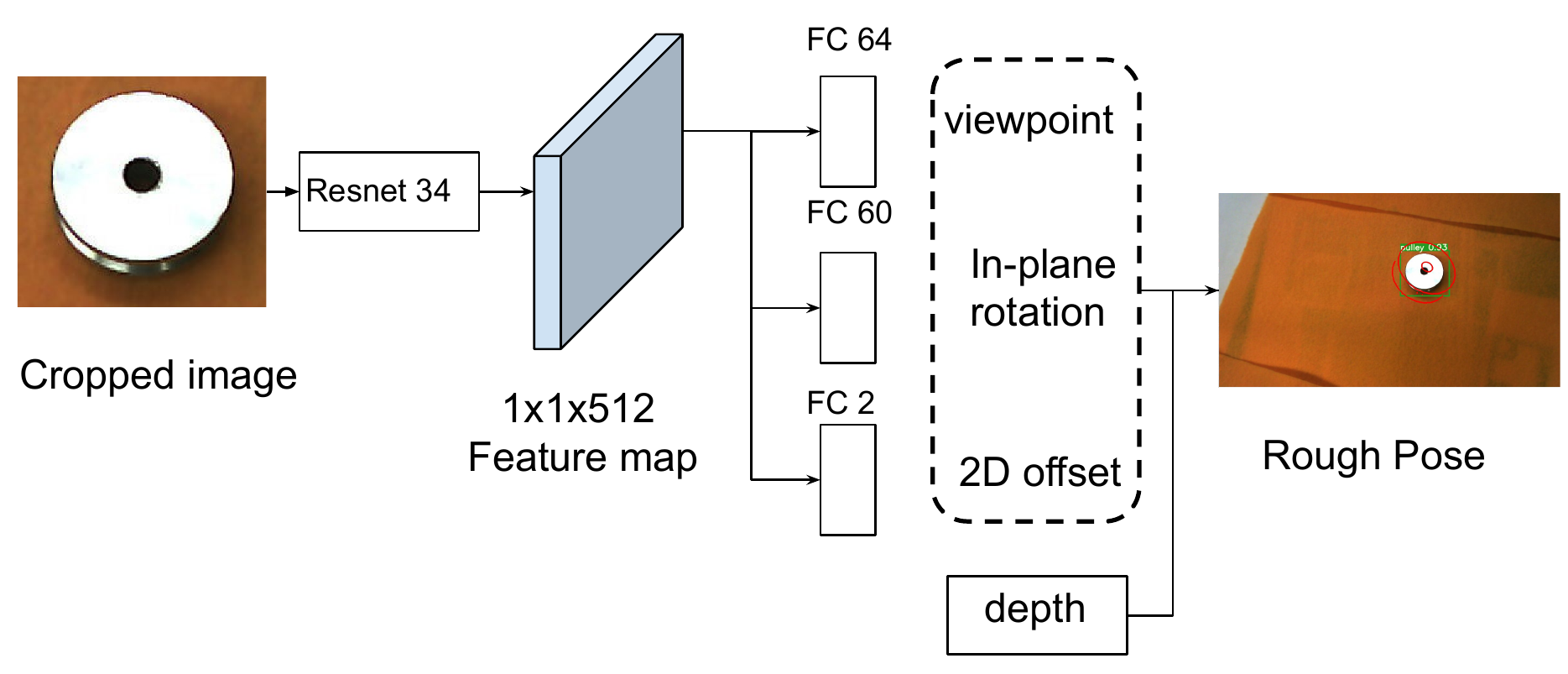}
\caption{Rough pose estimation network architecture.}
\label{rough}
\end{figure}

\subsubsection{Loss function} 
In our work, the loss function is built to include viewpoint ($L_{vp}$), and in-plane rotation ($L_{ipr}$), and 2d offset ($L_{2Doff}$) into account. Compared to SSD6D\cite{kehl2017ssd}, we do not include the class classification because object detection of the first stage will handle it. This is another instance of how using multiple stages helps us tackle a complex problem with an aggregate of simpler solutions. Here is our loss function:\\
\[
Loss = L_{vp} + L_{ipr} + L_{2Doff}
\]
In our loss functions, we consider $L_{vp}$ and $L_{ipr}$ as classification terms, and we employ a cross-entropy loss on them. On the other hand, we consider $L_{2Doff}$ as a regression term and employ the MSE loss on it. 
\begin{figure*}[!t]
\centering
\includegraphics[width=6.5in]{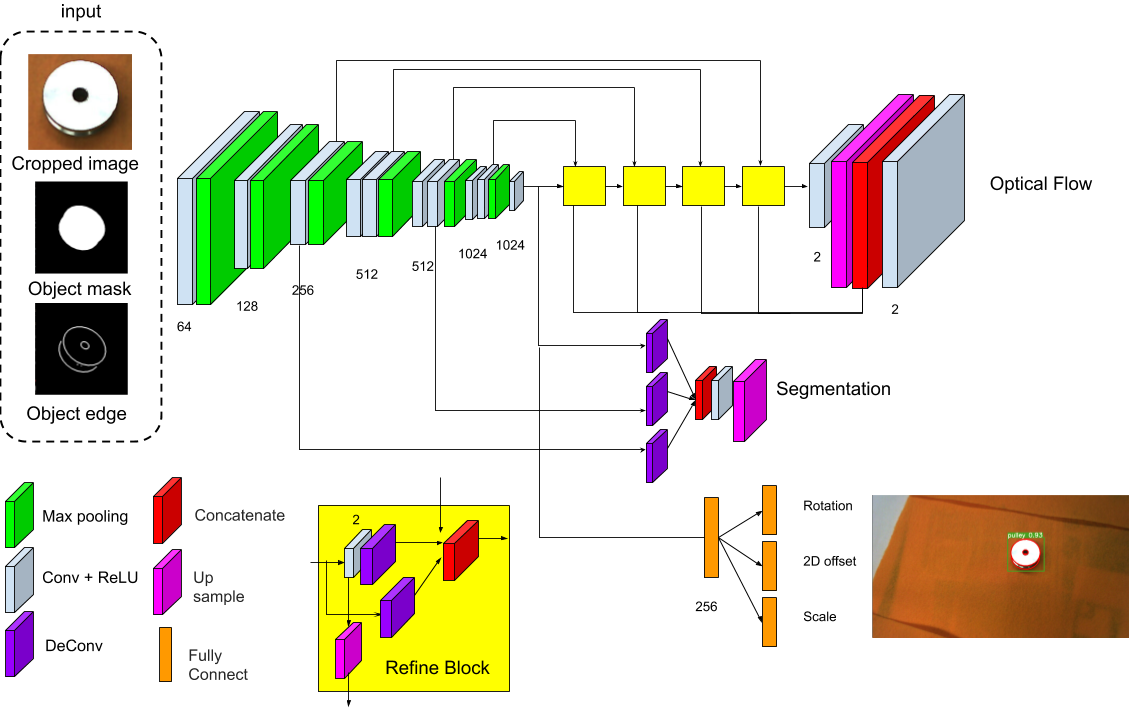}
\caption{Architecture of pose refinement}
\label{refine}
\end{figure*}

\subsection{Pose refinement}

\subsubsection{Architecture detail}
Given an input image with the object mesh and the initial pose, pose refinement will predict a relative transformation that can be applied to the initial pose for improving accuracy. In current works \cite{li2018deepim,manhardt2018deep}, instead of passing the initial pose and input image to the network directly, the initial pose with object mesh and material files is usually used to render the object, and the rendered object as an RGB image and input image will be given to the network which predicts the transformation from the initial pose to the predicted pose. In those methods, two assumptions exist. One is that the surface information of the object is required, and the other is that the surface of the object should not vary too much under different light conditions and viewpoints. However, specular and textureless objects break both those assumptions.

For specular objects, edges between two adjacent faces whose normals are close to perpendicular are more robust compared to other features, and those edges are called “sharp edge” \cite{choi2012robust}. That is, even though the face color of a specular object always reflects the color from the environment and changes frequently, most of the sharp edges are still visible in images. We propose to render sharp edges and semantic mask of the object as the input of our network, therefore using specular object specific information to aide training with modelled data.

In our work, given an initial pose and object mesh file, we render the object mask and sharp edges with OpenGL. With the RGB image, we crop the object from the input image and resize it to a specific size (we select 240 x 240 as the smallest crop being large enough to accommodate the largest object and the background detail). After that, the cropped image will be concatenated with the object edges and mask as a five-channel tensor input(3 channels for image, 1 channel for edge image, and 1 channel for segmentation mask) as an input of the network. For the network architecture, the FlowNetSimple will be used as the backbone. As shown in Figure \ref{refine}, the output of the network has three branches. One is used to predict the delta rotation, 2D offsets and changed scale. With those outputs, the transformation from initial pose to ground truth pose can be calculated. Moreover, in our study, we recognized that training with the optical flow and segmentation masks will boost the pose refinement performance, so there are two more branches after the FlowNetSimple for predicting the optical flow and segmentation of the object. We provide the reasoning of this improvement and the experiment result in the Ablation study in Section \ref{sec:ablation}.

\subsubsection{Loss functions}
The simple way to train the pose estimation network is to use the sum of the rotation loss and translation loss. However, there are multiple studies \cite{li2018deepim, kendall2017geometric} mentioning that using two different losses for pose estimation will suffer the problem of balancing them. In this section, we will demonstrate a common loss function(3D points distance, $L_{3Dpm}$) and explain its pros and cons, then argue why our new loss function(Cosine, $L_{cpm}$) is better than it for 6D poses.

3D point matching loss is calculating the average L1 distance between 3D points in the scene using the ground truth pose and the estimated pose. Given the predicted pose $P'=\left[R'| T'\right]$ and ground truth pose $P=\left[R| T\right]$, for each sample point $x_{i}$ on the object model, we have ground truth position $\left( Rx_{i}+T\right)$ and predict position $\left( R'x_{i}+T'\right)$. Thus, the 3D point marching loss will be
\[
L_{3Dpm}(P,P') = \frac{1}{n} \sum_{i = 1}^n L_1((Rx_i + T) - (R'x_i + T'))
\]
where n is the number of points, $L_1$ function is L1 norm. Essentially, the 3D point matching loss function measures how the real 3D model mesh against the 3d  model mesh with predicted pose, so balancing the weight on the loss of translation and the loss of rotation is not required. However, in our experiment, the loss caused by the translation is usually larger than the loss caused by the rotation, so the network will prefer to improve the prediction on the translation instead of rotation.

In our paper, we propose a loss function to prefer updating the rotation and call it “cosine point matching loss”. Instead of measuring the distance between positions of each point with predicted pose and ground truth pose, the cosine point matching is trying to measure the direction error of each sample point. Given the predicted pose $P'=\left[R'| T'\right]$ and ground truth pose $P=\left[R|T\right]$, for each sample point $x_{i}$ on the object model, we have ground truth position $\left( Rx_{i}+T\right)$  and predict position $\left( R'x_{i}+T'\right)$. Suppose the center point of object is $P_{center}$, the position of object center in camera frame is $\left( RP_{center}+T\right)$. Then, we can get two vectors V = $\left( \left(Rx_{i}+T\right)-\left( RP_{center}+T\right)\right)$ and V’ = $\left( \left(R'x_{i}+T'\right)-\left( RP_{center}+T\right)\right)$ for each sample point. The loss of each sample point is calculated as the negative cosine similarity between two vectors $-\frac{V\cdot V'}{\left\| V\right\| \times \left\| V'\right\| }$.
\[
L_{cpm}(P,P') = \frac{1}{n} \sum_{i = 1}^n -\frac{V\cdot V'}{||V||\times ||V'||}
\]
where $n$ is the number of points. 
Within this loss function, the loss caused by translation becomes nonlinear, so the loss ratio caused by translation error will be lower compared to 3d point matching loss function. 

\section{Implementation detail}
\subsection{Data collection and labeling}
\label{sec:datacollection}
There are two popular methods for collecting RGB-based image-pose datasets: video with object tracking and images of objects surrounded with QR codes. While the latter gives better accuracy the codes can confound object detection networks as unintended features of the image. Former is preferred for detection networks but is prone to errors in tracking. We designed our pipeline to divide object detection and pose estimation into two different parts, significantly this choice helps us in using the more accurate method for collecting pose-labeled dataset. We collect one dataset without QR codes for object detection network and one with codes for pose-estimation network. The codes are used for pose labeling but then the objects are cropped out of the image based on bounding box provided by the detection network.

\begin{figure}[!t]
\centering
\includegraphics[width=3.5in]{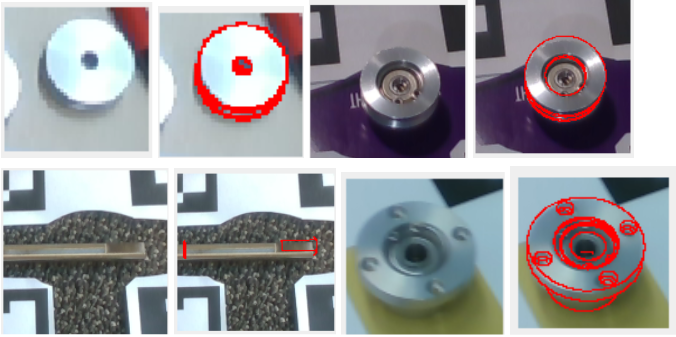}
\caption{All objects in our dataset}
\label{objects}
\end{figure}

\subsection{Training order}
In our experiment, directly training the refinement network will take a long time, so we pre-train only object optical flow and object mask at the beginning. After the IOU(Intersection over union) of the prediction on the semantic mask arrives 85$\%$, we can then train the network to predict pose transformation with optical flow and object mask.

\subsection{Compare pose regression}
In the second stage, we use a pose classifier to predict the viewpoints and in-plane rotation. Compared to the pose classifier, the pose regression which directly does rotation regression and offset regression is also popular in pose estimation like PoseNet. However, regression will be unreliable when it tries to predict the pose of a symmetric object. In the paper\cite{kehl2017ssd}, they also mentioned that the layers will have better performance at scoring discrete viewpoints than at outputting numerically accurate translation and rotations. Moreover, symmetric objects have similar look in different views, so it will be bias to predict proper pose with pose regression. On the other hand, compared to pose regression, viewpoint-based methods can easily solve the symmetric problem by mapping several poses with similar view into one pose.

\section{Experiment}
This section presents performance and evaluation of the proposed approach. Section \ref{sec:eval} introduces the metrics used in the experiment. Section \ref{sec:result} shows the results of the approach on our dataset. Section \ref{sec:ablation} presents an ablation study about the effectiveness of each components.
\subsection{ Evaluation metrics}
\label{sec:eval}
We use the following evaluation metrics for 6D object pose estimation. Average Distance metric as proposed in \cite{xiang2017posecnn} (ADD) computes the average pairwise distance between the 3D model points transformed using the predicted pose $P'=\left[R'| T'\right]$ and the ground truth pose $P=\left[R| T\right]$ :
\[
ADD = \frac{1}{n} \sum_{x \in M} ||(Rx + T) - (R'x + T') ||
\]
Where M denotes the set of sampled 3D model points and n is the total number of sampled points. The predicted pose is considered to be correct if the average distance is smaller than a predefined threshold. In our experiment, we choose to use 10$\%$ of the diameter to be the threshold.

Because objects in our domain have significant symmetries and ADD does not work well on symmetric objects, ADD-S\cite{xiang2017posecnn} is used which uses closest point distance in calculating the average distance:
\[
ADD-S = \frac{1}{n} \sum_{x_1 \in M} \min_{x_2 \in M} ||(Rx_1 + T) - (R'x_2 + T') ||
\]
Where M denotes the set of sampled 3D model points and n is the total number of sampled points. Similar to ADD, the predicted pose is considered to be correct if the average distance is smaller than a predefined threshold. In our experiment, we choose to use 1$\%$ of the diameter to be the threshold.

\begin{table}[!b]
\renewcommand{\arraystretch}{1.3}
\caption{Percentage of correct pose estimations using ADD with 10\% of diameter as success threshold and ADD-S with 1\% of the diameter as success threshold.}
\label{result}
\centering
\begin{tabular}{|c|c|c|}
\hline
 \textbf{Object} & \textbf{ADD} & \textbf{ADD-S}\\
\hline
Pulley & 52 & 95 \\
\hline
Housing & 87.6 & 99 \\
\hline
Nut & 2.2 & 61 \\
\hline
Shaft & 25.1 & 89.8\\
\hline
Pulley with Screw & 33.4 & 99\\
\hline
\end{tabular}
\end{table}

\begin{table*}[!thbp]
\renewcommand{\arraystretch}{1.3}
\caption{Ablation study of training model with optical flow and segmentation. Metric used is ADD with 10\% of diameter as threshold}
\label{component}
\centering
\begin{tabular}{|c|c|c|c|c|}
\hline
 \textbf{Object} & \textbf{Original model} & \textbf{Original model - Optical flow} & \textbf{Original model - Segmentation} & \textbf{Original model - Both}\\
\hline
Pulley & 49.52 & 45.78 & 49.02 & 46.62 \\
\hline
Housing & 62.38 & 55.92 & 61.36 & 59.61 \\
\hline
Shaft & 21.31 & 17.64 & 20.01 & 15.10\\
\hline
Pulley with screw & 29.69 & 29.07 & 27.92 & 17.10\\
\hline
\end{tabular}
\end{table*}

\subsection{Result on our dataset}
\label{sec:result}
\begin{figure}[!t]
\centering
\includegraphics[width=1.0in]{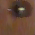}
\caption{Bad image in the dataset.}
\label{fail}
\end{figure}

In our dataset, there are five specular and symmetric objects, and four of them except Nut are shown in the figure \ref{objects}. In our evaluation process, we calculate ADD score and ADD-S score with the best hyper-parameters we found. As the Table \ref{result} shows, the performance of our pipeline on the dataset shows that pose estimation on specular objects with radial or bidirectional symmetry is feasible with our approach. However, the performance on Nut is poor. With our analysis, such poor performance is caused by the low resolution of the dataset. Because the size of nut is small, the cropped image of nut will be resized to a larger size, so the input image of the refinement module will be blurred like figure \ref{fail}. We designed our dataset collection parameters with the pulley, housing and shaft first and then extended it to the nut. The difference in object size distribution, and the unsatisfactory upper limit on resolution with our camera (Intel Real Sense) contributed to this dataset quality problem.

\begin{figure}[!b]
\centering
\includegraphics[width=2.5in]{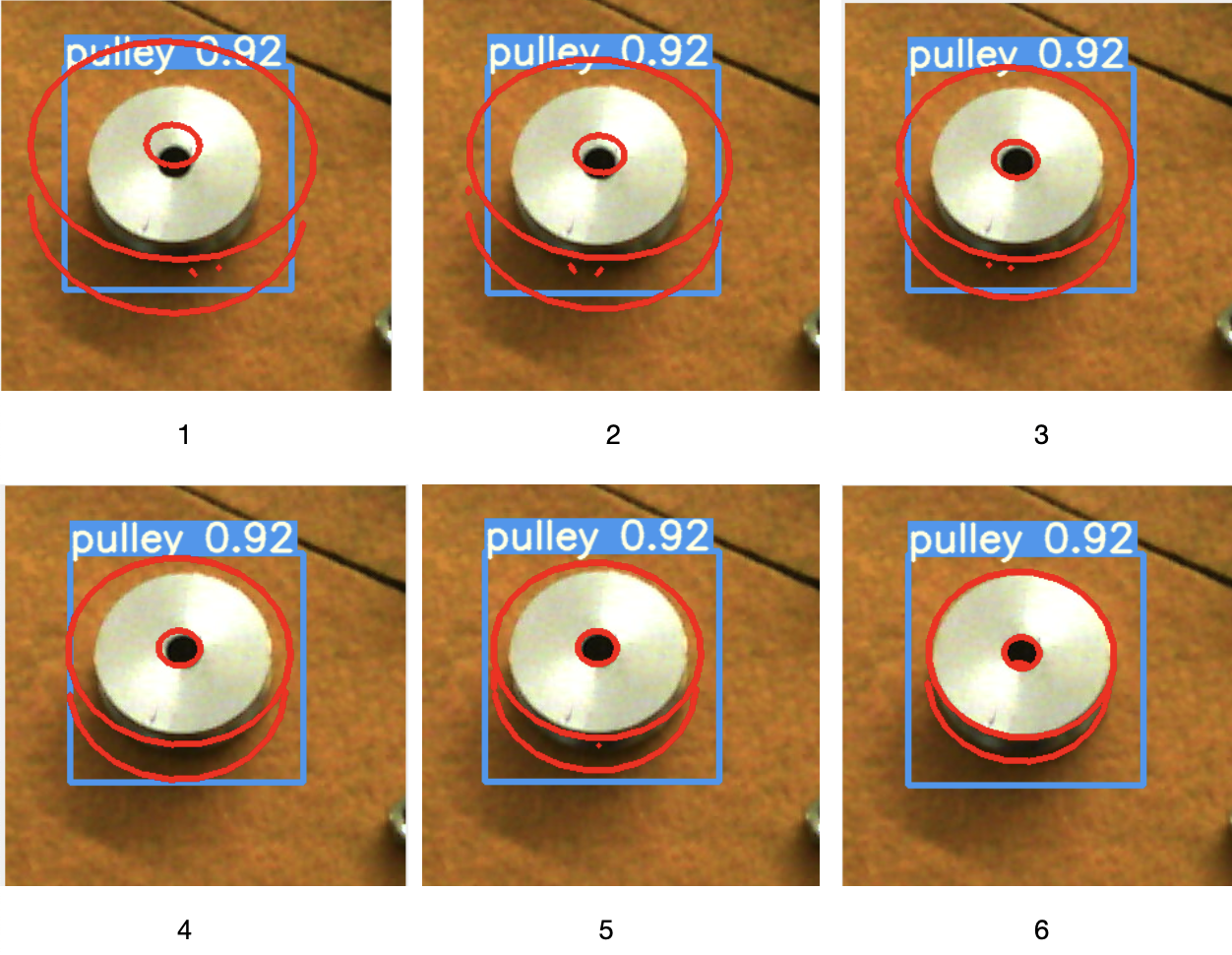}
\caption{Result of iterative refinement}
\label{iteration}
\end{figure}

In our experiment, single one step refinement in our architecture does not always give the correct pose with high accuracy but we observed if we repeat refinement with the last refined pose the output approaches higher accuracy. Even DeepIM \cite{li2018deepim} suggests that iteratively re-rendering the object based solution can improve pose estimation because the rendered image and the RGB image become more and more similar. Therefore, we design to use iterative training and iterative testing in our work. That is, the predicted pose will be used as initial pose several times for pose refinement during both training and testing. This is shown in figure \ref{iteration} shows the accuracy from iteratively refined pose of objects.

\begin{table}[!b]
\renewcommand{\arraystretch}{1.3}
\caption{Ablation study of different Loss function with ADD metric}
\label{loss}
\centering
\begin{tabular}{|c|c|c|}
\hline
 \textbf{Object} & \textbf{L1 loss} & \textbf{Cosine}\\
\hline
Pulley & 47.75 & 49.52 \\
\hline
Housing & 59.61 & 62.38 \\
\hline
Nut & 3.13 & 3.38 \\
\hline
Shaft & 16.85 & 21.31\\
\hline
Pulley with Screw & 29.55 & 29.68\\
\hline
\end{tabular}
\end{table}

\subsection{Ablation study}
\label{sec:ablation}
In this section, the effectiveness of different components of refinement module in our approach is measured with corresponding experiments. To simplify the study, we manually generate initial pose with noise from the label pose instead of using rough pose estimation module. We rotate the object with a randomly sampled degree and translate the object with a randomly generated offset. The degree is drawn from a normal distribution having a mean of 0.0 and variance of 0.3, while the offset is drawn from a normal distribution having a mean of 0.0 and variance of 0.01, and the depth is drawn from a normal distribution having a mean of 0.0 and variance of 0.08.

\subsubsection{Training model with optical flow and segmentation}
Table \ref{component} shows the results that training model with and without optical flow and segmentation. The metric using in this table is ADD rate. We can see that training model with both optical flow and segmentation could increase the performance a lot. We believe that the reason is such training strategy could make the network to focus on the object and learn rotation information. 

\subsubsection{Loss function}
Table \ref{loss} shows the results of using different loss functions. According to the table, the Cosine loss function is measurably better than the 3D points matching in most of cases. However, since it is more complex the cosine point matching loss also requires generous sample points on the object. In our experiment, when the number of sample points is low, model using cosine loss function performs even worse than the one using 3D points matching loss. Compared to cosine point matching loss, 3D points matching loss function is not sensitive to number of sample points and the performance is stable with different number of sample points.

\section{Conclusion}
In this work, we propose a cascaded neural network architecture based 6D pose estimation approach which can deal with symmetric problems and does not require the surface detail of objects. Our approach break this complex problem into several small problems: object detection, coarse pose estimation, pose refinement. Given an RGB image, our approach use YOLO to detect the object. A coarse pose is calculated by a rough pose estimator and finally pose refinement could estimate an accurate pose for picking it up. Furthermore, we introduce a new loss function (cosine point matching loss) which could improve the performance a lot. On our dataset, it outperforms the 3D points matching loss. 

This work has proven that 6D pose estimation could be done without texture information and we use sharp edge image as a feature of a pose. It is interesting to explore other feature of the pose of a textureless object. 

\section*{Acknowledgement}
This research has been sponsored by RPDC (Contact +966 (11) 221 1111). We much appreciate the collaboration with Dr. Nahid Sidki, Dr. Omar Aldughayem, and Hussam Alzahrani.




%

  

{
\bibliographystyle{IEEEtran}
\bibliography{egbib}
}

\end{document}